\documentclass[11pt]{article}
\usepackage{coling2020}
\usepackage{times}
\usepackage{url}
\usepackage{latexsym}

\usepackage{microtype}
\usepackage{latexsym}
\usepackage{array}
\usepackage{color}
\usepackage{bm}
\usepackage{enumitem}
\usepackage{booktabs}
\usepackage{amssymb}
\usepackage{amsfonts}
\usepackage{multirow}
\usepackage{subcaption}
\usepackage{graphicx}
\usepackage{amsfonts}
\usepackage{amsmath}
\usepackage{CJKutf8}

\newcommand{\mask}{\texttt{[MASK]}}

\newcommand{\ourresult}{$ {}^{\dagger} $}
\newcommand{\ensemble}{$ {}^{*} $}
\newcommand{\baseline}{\textsc{MP-MLM}}
\newcommand{\fscore}{\ensuremath{\mathrm{F}_{1}}}

\newcommand{\alns}[1] {\begin{align} #1 \end{align}}
\newcommand{\mathcenter}[1]{\begin{center} \vspace{-0.4cm} #1 \end{center}}

\newcommand{\masknoun}{$\mathcal{S}_{\mathrm{noun}}$}
\newcommand{\maskverb}{$\mathcal{S}_{\mathrm{verb}}$}
\newcommand{\maskparticle}{$\mathcal{S}_{\mathrm{particle}}$ }
\newcommand{\masksymbol}{$\mathcal{S}_{\mathrm{symbol}}$}
\newcommand{\masknonoun}{$\mathcal{S}_{all\backslash\mathrm{noun}}$}
\newcommand{\masknoverb}{$\mathcal{S}_{all\backslash\mathrm{verb}}$}
\newcommand{\masknoparticle}{$\mathcal{S}_{all\backslash\mathrm{particle}}$}
\newcommand{\masknosymbol}{$\mathcal{S}_{all\backslash\mathrm{symbol}}$}
\newcommand{\masknoverbnosymbol}{$\mathcal{S}_{all\backslash(\mathrm{verb} \cup \mathrm{symbol})}$}
\newcommand{\maskall}{$\mathcal{S}_{\mathrm{all}}$}

\DeclareMathOperator*{\argmax}{arg\,max}

\DeclareMathOperator*{\MLM}{\mathrm{MLM}}

\makeatletter
\newcommand{\figcaption}[1]{\def\@captype{figure}\caption{#1}}
\newcommand{\tblcaption}[1]{\def\@captype{table}\caption{#1}}
\makeatother

\newcommand{\todo}[1]{}
\renewcommand{\todo}[1]{{\color{red} TODO: {#1}}}

\colingfinalcopy 

\title{An Empirical Study of Contextual Data Augmentation for \\ Japanese Zero Anaphora Resolution}

\author{
  \bf{Ryuto Konno}$^{\,\spadesuit}$ ~~
  \bf{Yuichiroh Matsubayashi}$^{\,\spadesuit\diamondsuit}$ ~~
  \bf{Shun Kiyono}$^{\,\diamondsuit\spadesuit}$ \\
  \bf{Hiroki Ouchi}$^{\,\diamondsuit}$ ~~
  \bf{Ryo Takahashi}$^{\,\spadesuit\diamondsuit}$ ~~
  \bf{Kentaro Inui}$^{\,\spadesuit\diamondsuit}$ \\
  ${}^{\spadesuit}$Tohoku University ~~ ${}^{\diamondsuit}$RIKEN \\
  {\tt \{ryuto, ryo.t, inui\}@ecei.tohoku.ac.jp} \\
  {\tt y.m@tohoku.ac.jp } \\
  {\tt \{shun.kiyono, hiroki.ouchi\}@riken.jp} \\
}

\date{}

\begin{document}
\maketitle
\begin{abstract}
One critical issue of zero anaphora resolution (ZAR) is the scarcity of labeled data.
This study explores how effectively this problem can be alleviated by data augmentation. 
We adopt a state-of-the-art data augmentation method, called the contextual data augmentation (CDA), that generates labeled training instances using a pretrained language model.
The CDA has been reported to work well for several other natural language processing tasks, including text classification and machine translation~\cite{kobayashi2018,wu2019,gao2019}. 
This study addresses two underexplored issues on CDA, that is, how to reduce the computational cost of data augmentation and how to ensure the quality of the generated data.
We also propose two methods to adapt CDA to ZAR: \textit{\mask{}-based augmentation} and \textit{linguistically-controlled masking}.
Consequently, the experimental results on Japanese ZAR show that our methods contribute to both the accuracy gain and the computation cost reduction.
Our closer analysis reveals that the proposed method can improve the quality of the augmented training data when compared to the conventional CDA.
\end{abstract}

\section{Introduction}
\label{sec:introduction}

\blfootnote{
    \hspace{-0.65cm}
    This work is licensed under a Creative Commons 
    Attribution 4.0 International License.
    License details:
    \url{http://creativecommons.org/licenses/by/4.0/}.
}

\begin{figure}[t]
    \centering
    \includegraphics[width=\hsize]{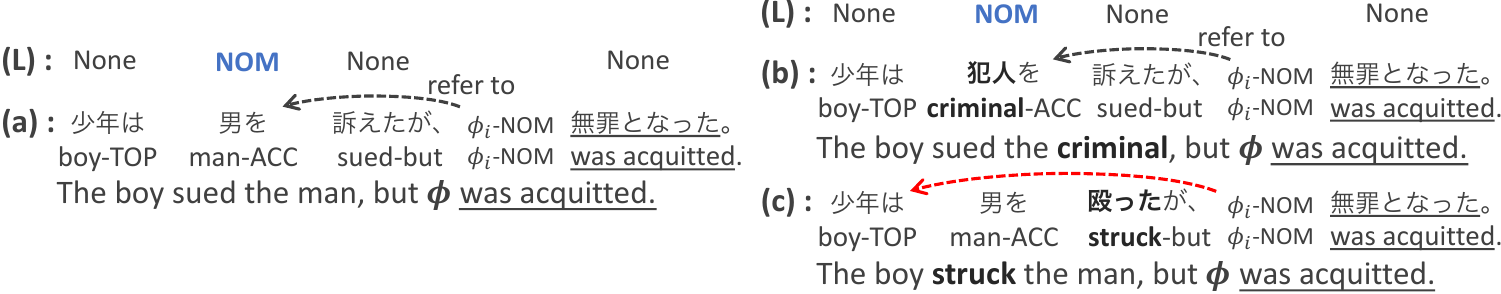}
    \caption{
        Examples of anaphoric relations in augmented sentences by CDA. ``(L)'': the gold labels. ``(a)'': the original sentence. ``(b)'': successfully augmented sentence. ``(c)'': undesirable augmented sentence.
    }
    \label{fig:replace-example}
\end{figure}

In pro-drop languages, such as Japanese and Chinese, the arguments of a predicate are frequently omitted in sentences.
Automatic recognition of such omitted arguments plays an important role in understanding the predicate-argument structure of a sentence.
This task is referred to as zero anaphora resolution (ZAR).
Figure~\ref{fig:replace-example}a depicts an example of the task.
In the sentence, the nominative argument of the predicate \textit{was acquitted} is omitted.
Such an omitted argument is referred to as a \textit{zero pronoun} often represented as $\phi$.
In Figure~\ref{fig:replace-example}a, $\phi$ refers to \textit{the man}.

One critical issue of ZAR is the scarcity of labeled data.
To compensate for the scarcity, several studies have explored the direction of exploiting unlabeled data~\cite{sasano-etal-2008-fully,sasano2011discriminative,chen2014chinese,chen2015chinese,liu:2017,yamashiro-etal-2018-neural,kurita2018}.
Another promising direction is to automatically augment labeled data (data augmentation).
One state-of-the-art data augmentation method is contextual data augmentation (CDA), which augments labeled data by replacing an arbitrary token(s) with another token(s) predicted to be plausible in the surrounding context by a pretrained language model (LM).
CDA works well in several natural language processing (NLP) tasks, such as text classification~\cite{kobayashi2018,wu2019} and machine translation~\cite{gao2019}.
However, unlike the direction of exploiting unlabeled data, the potential of data augmentation in the context of ZAR has largely remained underexplored.
This study investigates how the idea of CDA can be effectively employed in ZAR, taking the task of Japanese ZAR as a case study.

Figure~\ref{fig:replace-example} illustrates a straightforward approach of applying the idea of CDA to ZAR.
For each given training instance (i.e., input sentence (a) paired with its gold labels (L)), a standard CDA method would use a pretrained LM to generate a variant(s) of the original sentence, say sentence (b). The new sentence (b) is then paired with the original gold labels (L) and is added to the training set (detailed in Section~\ref{subsec:CDA}). 
This straightforward approach raises two critical issues. 
One is that the computational cost of CDA can be non-trivial because it runs a huge pretrained LM for each input sentence in the training set. 
Note that even with a relatively small training set, this cost can be enormous when one wants to repeatedly conduct experiments with a large variety of settings. 
The second issue is that CDA may produce improper instances. 
Figure~\ref{fig:replace-example}b shows that the original noun \textit{man} is properly replaced with another noun \textit{criminal}.
By saying that a word is properly replaced with another, we mean that the replacement does not affect the anaphoric relation labels of the original sentence.
This requirement is critical because we want to use produced instances with the original ZAR signals to train our model. 
However, if not appropriately controlled, CDA may disrupt the original anaphoric relations, as in Figure~\ref{fig:replace-example}c, where replacing the verb \textit{sued} with \textit{struck} affects the anaphoric relation between $\phi$ and \textit{boy}. 
These issues can be crucial when applying CDA to ZAR, but they have not been addressed in previous works with CDA for other NLP tasks. 

To address these two issues, we newly design a CDA variant with two directions of extension: (i) \textit{\mask{}-based augmentation} and (ii) \textit{linguistically-controlled masking}.
Unlike a standard CDA method, which replaces tokens with other tokens, the \mask{}-based augmentation replaces tokens with the \mask{} token (Section~\ref{sec:mask-based-augmentation}). 
This simple extension halves the cost of running a pretrained LM.
Our experiments show that it also improves the performance of ZAR.
The second extension, which is \textit{linguistically-controlled masking}, provides a means of controlling the types of tokens to be replaced to inhibit improper token replacements (Section~\ref{sec:linguistically-controlled masking}).
The token replacement quality must be considered even in the \mask{}-based augmentation scheme because replacing a token with the \mask{} token may also affect the anaphoric relations.
Our experiments demonstrate that controlling token replacement by part-of-speech (POS) tags effectively inhibits improper token replacements and improves the performance of ZAR (see Section~\ref{sec:analysis} for details). 

Through extensive experiments, we reveal the relationship between performance gain and replacement of each token type with various masking probabilities.
We deepen our understanding of the change of antecedents through a detailed analysis on actual augmented data.
In summary, our main contributions are as follows:
\begin{itemize}
\setlength{\parskip}{0cm} 
\setlength{\itemsep}{0.1cm}
\item Two technical modifications to a standard CDA method;
\item Extensive empirical results indicating which type of tokens should (or should not) be the target for replacement; and
\item In-depth analysis on the change of antecedents of zero pronouns, suggesting a direction for future improvements.
\end{itemize}

\section{Related Work}
\label{sec:related-work}
ZAR has been studied in the Asian languages, such as Chinese~\cite{converse-2005-resolving,zhao2007identification,kong-zhou-2010-tree,chen2016chinese,yin2018zero}, Japanese~\cite{seki-etal-2002-probabilistic,isozaki-hirao-2003-japanese,iida-etal-2003-incorporating,iida-etal-2006-exploiting,Iida2007,Iida2015,iida:2016:intra}, and Korean~\cite{han2006korean}, and Romance languages, such as Italian~\cite{rodriguez-etal-2010-anaphoric,iida-poesio-2011-cross} and Spanish~\cite{ferrandez-peral-2000-computational,palomar2001algorithm}.
While many recent studies adopted supervised learning methods,\footnote{Some unsupervised methods have been studied in Chinese ZAR~\cite{chen2014chinese,chen2015chinese}.} the amount of labeled data is not sufficient to teach a model about knowledge for accurately resolving anaphoric relations.
Some studies tried extracting and exploiting such knowledge from large-scale unlabeled data (e.g., case frame construction~\cite{sasano-etal-2008-fully,sasano2011discriminative,yamashiro-etal-2018-neural}, pseudo training data generation ~\cite{liu:2017}, and semi-supervised adversarial training~\cite{kurita2018}).
Although findings on using unlabeled data for ZAR have been accumulated, the automatic augmentation of labeled data has been underinvestigated.
One reason for this is the difficulty in producing high-quality pseudo data from its original labeled data.

Some existing methods of text data augmentation take advantage of hand-crafted resources~\cite{zhang2015character,wang:2015} and rules~\cite{furstenau:2009,kafle:2017}.
With the recent advances of LMs~\cite{peters:2018:elmo,devlin:2019:NAACL}, CDA, which is a new method using LMs, has been proposed and reported to effectively increase labeled data in several NLP tasks~\cite{kobayashi2018,wu2019,gao2019}.
This method offers a variety of good substitutes of original tokens without any hand-crafted resources.
In ZAR, a key to performance improvement is training a model on anaphoric relations in various contexts.
However, the original labeled data over a limited portion of contextual diversity.
Here, CDA is a promising tool for producing good contextual variations of each original sentence.
Considering this advantage, we introduce CDA to ZAR and provide extensive empirical results to encourage the future exploration of this direction.

\section{Japanese Zero Anaphora Resolution}
\label{sec:japanese-zar}

\subsection{Problem Formulation and Notation}
\label{subsec:problem-formulation}
Japanese ZAR is often formulated as a part of the predicate-argument structure (PAS) analysis, which includes the task of detecting syntactically depending arguments (DEPs), in addition to ZAR.
Given a sentence $\bm{X}$ and a list of predicates $\bm{p}$, the Japanese PAS analysis task aims to identify the head words of nominative (\texttt{NOM}), accusative (\texttt{ACC}), and dative (\texttt{DAT}) arguments for each predicate.
Our task definition strictly follows that of previous studies~\cite{Iida2015,ouchi2017,Matsubayashi2017,matsubayashi:2018:COLING}, making our results comparable with those published previously.

In the following sections we consider that the input sentence $\bm{X} = (\bm{x}_1,\dots,\bm{x}_I)$ consists of a sequence of one-hot vectors $\bm{x}_{i} \in \{0,1\}^{|\mathcal{V}|}$, each of which represents a word in the sentence.
$\mathcal{V}$ denotes a vocabulary set.
Each sentence comprises $J$ predicates $\bm{p} = (p_{1},\dots,p_{J})$, where $p_j \in \mathbb{N}$ is a natural number indicating a predicate position.

\subsection{Baseline Model}
\label{subsec:baseline-model}
\begin{figure}[t]
    \centering
    \begin{minipage}[b]{.39\linewidth}
        \centering
        \includegraphics[width=\hsize]{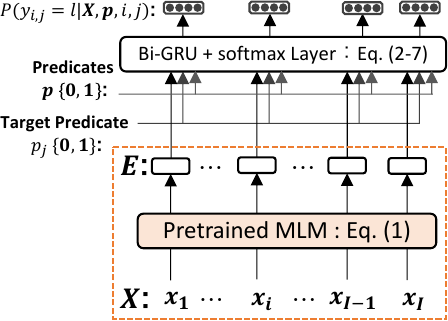}
        \subcaption{Baseline}
        \label{fig:baseline}
    \end{minipage}
    \hspace{.01\linewidth}
    \begin{minipage}[b]{.28\linewidth}
        \centering
        \includegraphics[height=\hsize]{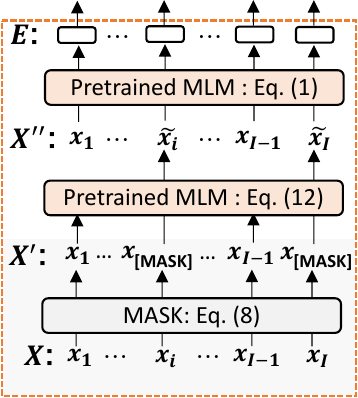}
        \subcaption{CDA}
        \label{fig:overview-CDA}
    \end{minipage}
    \hspace{.01\linewidth}
    \begin{minipage}[b]{.28\linewidth}
        \centering
        \includegraphics[height=\hsize]{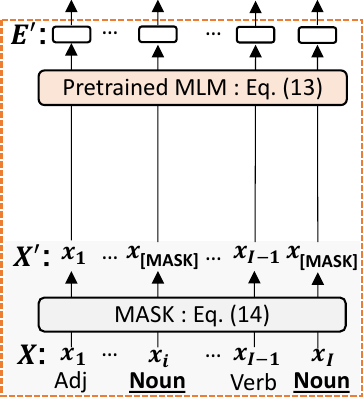}
        \subcaption{Masking}
        \label{fig:overview-masking}
    \end{minipage}
    \label{fig:overview-method}
    \caption{
        Overview of our models.
        (a): the baseline model (\baseline{}) for Japanese ZAR.
        (b): the integration of a standard CDA method into the baseline. 
        (c): proposed data augmentation method.}
\end{figure}

The baseline model employed in our experiments (hereinafter referred to as MP-masked language model (\baseline{})) is based on the multi-predicate (\textsc{MP}) model proposed by  Matsubayashi and Inui~\shortcite{matsubayashi:2018:COLING}.
Figure~\ref{fig:baseline} depicts an overview of the \baseline{}.
The only difference is that our \baseline{} uses a sequence of the final hidden states of a pretrained MLM, such as BERT~\cite{devlin:2019:NAACL}, as the input for the \textsc{MP} model, whereas
the original MP model uses conventional non-contextual word embeddings as the input layer.
The \baseline{} is a sequential labeling model.
Given a sentence $\bm{X}$, predicate positions $\bm{p}$ and a target predicate position $p_j \in \bm{p}$ formally.
The \baseline{} models a conditional probability $P({y}_{i,j} | \bm{X}, \bm{p}, i, j)$. 
Here, ${y}_{i,j} \in \{\texttt{NOM}, \texttt{ACC}, \texttt{DAT}, \texttt{NONE}\}$ is an argument label for a pair of $i$th word $\bm{x}_{i}$ and $j$th predicate.

The \baseline{} encodes an input $\bm{X}$ as follows:
\alns{\bm{E} = \MLM(\bm{X}),}
where $\MLM$ represents a function computing its final hidden states $\bm{E} = (\bm{e}_1,\dots,\bm{e}_I)$.\footnote{A given token $\bm{x}_{i}$ may be split into multiple subwords depending on the MLM vocabulary. Following the previous studies~\cite{devlin:2019:NAACL,kondratyuk:2019:EMNLP}, all subwords in the input sentence were fed into the MLM, and only the representation of the head subword was used for each token.}
Here, $\bm{e}_{i} \in \mathbb{R}^{D}$, and $D$ is the hidden state size.
Each state $\bm{e}_{i}$ is then concatenated with two binary values $b^{target}_i$ and $b^{others}_i$, where $b^{target}_i$ represents whether or not the word is the target predicate, and $b^{others}_i$ represents whether or not the word is the predicate.
This concatenation operation is presented as follows:
\alns{\bm{h}^{0}_{i,j} = \bm{e}_i \oplus b^{\mathrm{target}}_i \oplus b^{\mathrm{others}}_i,}
\mathcenter{
    \begin{minipage}[b]{.45\linewidth}
        \alns{
          b^{\mathrm{target}}_i =&
          \begin{cases}
            1 & (\mbox{if}\quad i = p_j) \\
            0 & (\mbox{otherwise}),
          \end{cases}
        }
    \end{minipage}
    \hspace{.05\linewidth}
    \begin{minipage}[b]{.48\linewidth}
        \alns{
          b^{\mathrm{others}}_i =&
          \begin{cases}
            1 & (\mbox{if}\quad i \in \bm{p}) \\
            0 & (\mbox{otherwise}),
          \end{cases}
        }
    \end{minipage}
}
where $\oplus$ denotes the concatenation operation, and $\bm{h}^{0}_{i,j} \in \mathbb{R}^{D+2}$ denotes the output of this concatenation operation.
The obtained vectors are then fed into the $k$-layer bi-directional RNN ($\textrm{BiRNN}$) with residual connections~\cite{He2016} and alternating directions~\cite{Zhou2015} as follows:
\alns{
  \bm{h}^1_{i,j} =& \textrm{RNN}^{1}(\bm{h}^{0}_{i,j}, \bm{h}^{1}_{i-1,j}), 
}
\alns{
  \bm{h}^k_{i,j} =&
  \begin{cases}
    \bm{h}^{k-1}_{i,j} + \textrm{RNN}^{k}(\bm{h}^{k-1}_{i,j}, \bm{h}^{k}_{i-1,j}) & (k \mbox{ is odd}) \\
    \bm{h}^{k-1}_{i,j} + \textrm{RNN}^{k}(\bm{h}^{k-1}_{i,j}, \bm{h}^{k}_{i+1,j}) & (k \mbox{ is even})
  \end{cases} (k \geq 2),
}
where $\bm{h}^{k}_{i,j} \in \mathbb{R}^{M}$ denotes the output of the $k$th RNN layer, and $\textrm{RNN}^{k}$ denotes the function representing the $k$th RNN layer.
Gated recurrent units~\cite{Cho2014} are used for RNN cells.
A four-dimensional vector representing a probability distribution $P({y}_{i,j}| \bm{X}, \bm{p}, i, j) \in \mathbb{R}^{4}$ is then obtained as follows by applying a softmax layer to the output $\bm{h}^K_{i,j} \in \mathbb{R}^{M}$:
\alns{P({y}_{i,j}| \bm{X}, \bm{p}, i, j) = \mathrm{softmax}(\bm{W} \bm{h}^K_{i,j}),}
where $\bm{W} \in \mathbb{R}^{4 \times M}$ is a classification layer.
For each argument slot $l \in  \{\texttt{NOM}, \texttt{ACC}, \texttt{DAT}\}$ of each predicate, we eventually select a word $\bm{x}_i$ with the maximum probability of $P({y}_{i,j}=l| \bm{X}, \bm{p}, i, j)$ as an argument if the probability exceeds a threshold $\theta_{l} \in \mathbb{R}$; otherwise, the slot remains empty.

\section{Method}
\label{sec:proposed-method}
In this section, we first review an existing standard CDA method (Section~\ref{subsec:CDA}), which is our starting point of data augmentation for ZAR.
We then discuss the \textbf{quantitative} and \textbf{qualitative} aspects of CDA: (i) \textit{computational inefficiency of symbolic replacement} and (ii) \textit{change of antecedents}.
Considering these aspects, we improve a standard CDA method with two methods, namely \textit{\mask{}-based augmentation} (Section~\ref{sec:mask-based-augmentation}) and \textit{linguistically-controlled masking} (Section~\ref{sec:linguistically-controlled masking}).

\subsection{Contextual Data Augmentation (CDA)}
\label{subsec:CDA}
The idea underlying CDA aims to replace token(s) in a given sentence with other token(s) that seem probable in a given context.
Sentences that contain some replaced tokens are regarded as new training instances.
The alternative tokens for the replacement are selected on the basis of the probability distribution output from a pretrained LM.
For example, in existing studies, bi-directional RNN-based LMs~\cite{kobayashi2018} and MLM~\cite{wu2019} have been used to compute the distribution.
Our CDA formulation closely follows that proposed by Wu et al.~\shortcite{wu2019}.
We first replace token(s) in $\bm{X}$ with \mask{} and obtain $\bm{X}^{\prime}$ as follows:
\mathcenter{
    \begin{minipage}[]{.67\linewidth}
        \alns{
            \bm{X}^{\prime} = (\bm{x}_{1}^{\prime},\dots,\bm{x}_{I}^{\prime})
            \quad \mbox{where} \quad
            \bm{x}^{\prime}_{i} =&
            \begin{cases}
            \bm{x}_{i} & (r_i = 0) \\
            \bm{x}_{\mask{}}   & (r_i = 1),
            \end{cases} \label{eqn:mask}
        }
    \end{minipage}
    \hspace{.02\linewidth}
    \begin{minipage}[]{.29\linewidth}
        \alns{
            r_i \sim& \mathrm{Bernoulli(\alpha)},
        }
    \end{minipage}
}
where $\alpha$ is the probability of the replacement with \mask{}.
We then feed $\bm{X}^{\prime}$ to MLM to obtain a sequence of the final hidden states $(\bm{e}^{\prime}_1,\dots,\bm{e}^{\prime}_I)$, where $\bm{e}^{\prime}_{i} \in \mathbb{R}^{D}$.
The MLM then computes the token for replacement $\tilde{\bm{x}}_i$ as
\mathcenter{
    \begin{minipage}[b]{.45\linewidth}
        \alns{P_i &= \mathrm{softmax}(\bm{W}_{\textrm{MLM}}\bm{e}^\prime{}_i),}
    \end{minipage}
    \hspace{.05\linewidth}
    \begin{minipage}[b]{.48\linewidth}
        \alns{\tilde{\bm{x}}_i &= \mathrm{F}(P_i), \label{eqn:sampling}}
    \end{minipage}
}
where $\bm{W}_{\textrm{MLM}} \in \mathbb{R}^{\lvert V \rvert \times D}$ denotes a classification layer of the pretrained MLM; $P_i \in \mathbb{R}^{\lvert V \rvert}$ denotes the probability distribution over the vocabulary; and $\mathrm{F}$ denotes a function for selecting one word from $P_i$.
\mask{}s are then replaced with the predicted tokens to obtain a new sequence $\bm{X}^{\prime\prime}$ as follows:
\alns{
  \bm{X}^{\prime\prime}  = (\bm{x}_{1}^{\prime\prime},\dots,\bm{x}_{I}^{\prime\prime})
  \quad \mbox{where} \quad \bm{x}_{i}^{\prime\prime} &=
  \begin{cases}
    \bm{x}_{i} & (\mbox{if}\quad\bm{x}_{i}^{\prime} \neq \bm{x}_{\mask{}}) \\
    \tilde{\bm{x}}_i  & (\mbox{otherwise}),
  \end{cases}
}
Finally, $\bm{X}^{\prime\prime}$ is paired with the original gold labels of $\bm{X}$ and regarded as a new training instance.

As we mentioned in Section~\ref{sec:introduction}, we improved the above CDA method.
First, we improved the \textbf{computational efficiency}. 
While good substitutes of the original tokens can be produced by using a gigantic MLM, the computational cost for training the model increases, and this cost cannot be ignored.
Although the amount of training data of the Japanese ZAR is relatively small, the computational cost can be enormous when one wants to conduct multiple experiments with various settings.
In Section~\ref{sec:experiment}, we indeed conduct extensive experiments to explore the use of CDA for ZAR.
We design a technique that halves the computational cost of incorporating the abovementioned CDA method into the baseline model.
This technique improves the performance of ZAR more effectively than a standard CDA on the same amount of augmented training data.
Second, we improved the \textbf{controllability}.
Our preliminary experiments suggested that replacing certain types of tokens is likely to change the antecedent of each zero pronoun.
If another token (a token different from the original one) can be interpreted as the antecedent after the replacement, the original label will make no sense, which leads to a model being confused.
We design herein a method that enables controlling the types of tokens to be replaced (Section~\ref{sec:linguistically-controlled masking}).

\subsection{\mask{}-based Augmentation}
\label{sec:mask-based-augmentation}
We present the \textit{\mask{}-based augmentation} and simplify the integration of the original CDA into the baseline model (Section~\ref{subsec:baseline-model}) for computational efficiency.
Figure~\ref{fig:overview-masking} shows an overview of the method.
Instead of using the sequence $\bm{X}^{\prime\prime}$ with token-replacement, we used the masked sequence $\bm{X}^{\prime}$ (Equation~\ref{eqn:mask}) as the new training data.
That is, while an MLM has to be run over each input sequence twice (for predicting substitutes and computing their feature representations) when using the original CDA, the MLM has to be run only once in our simplified one.
This one-pass running of the MLM halves the computational cost of the original CDA.
In particular, the masked sequence $\bm{X}^{\prime}$ is fed into the MLM, and a sequence of the final hidden states of MLM $\bm{E}^{\prime}$ is obtained:
\alns{
  \bm{E}^{\prime} &= \MLM(\bm{X}^{\prime}). \label{equation:augmented-bert}
}
The remaining computations are the same as those for the MP-MLM described in Section~\ref{subsec:baseline-model}.
Note that the MLM is pretrained to fill the \mask{} with a token that is probable in a given context. 
We can expect that the final hidden state of the \mask{} is an abstract and a mixture of contextually possible word representations for a given context.
This is not the case for the original CDA approach (Figure~\ref{fig:overview-CDA}), where the tokens are symbolically replaced with other tokens.

\subsection{Linguistically-controlled Masking}
\label{sec:linguistically-controlled masking}
We introduce and add a new function, called \textit{linguistically-controlled masking}, to the abovementioned framework \textit{\mask{}-based augmentation}.
This function enables us to freely choose which types of tokens will be the target for replacement.
As explained in Figure~\ref{fig:replace-example} in Section~\ref{sec:introduction}, replacing certain types of tokens is likely to change the antecedent of each zero pronoun.
Here, instead of replacing arbitrary tokens, we utilize POS tags to control the types of tokens to be replaced.
We particularly add the POS constraint to Equation~\ref{eqn:mask}:
\alns{
    \bm{X}^{\prime} = (\bm{x}_{1}^{\prime},\dots,\bm{x}_{I}^{\prime})
    \quad \mbox{where} \quad
    \bm{x}^{\prime}_{i} =&
    \begin{cases}
    \bm{x}_{i} & (r_i = 0) \\
    \bm{x}_{\mask{}}   & (r_i = 1 \quad\mbox{and}\quad s_i \in \mathcal{S}),
    \end{cases}
}
\mathcenter{
    \vspace{-0.2cm}
    \begin{minipage}[b]{.48\linewidth}
        \alns{r_i \sim& \mathrm{Bernoulli(\alpha)},}
    \end{minipage}
    \hspace{.015\linewidth}
    \begin{minipage}[b]{.49\linewidth}
        \alns{s_i =& \psi(\bm{X},i),}
    \end{minipage}
}
where $\mathcal{S}$ denotes a set of target POS tags to be replaced, and $\psi$ denotes a function for obtaining the POS tag $s_i$ of the $i$th word $\bm{x}_i$.
In other words, $\bm{x}_i$ is replaced with \mask{} only if its POS tag belongs to the set of target POS tags to be replaced.
The remaining operations are the same as those for the \baseline{} described in Section~\ref{subsec:baseline-model}.

\section{Experiments}
\label{sec:experiment}
\begin{table}[t]
    \small
    \begin{minipage}[b]{.48\linewidth}
        \centering
        \begin{tabular}{llrrr} 
            \toprule
            Split     &      & NOM    & ACC    & DAT \\ 
            \midrule
            Training   & DEP  & 36,877 & 24,624 & 5,741 \\
                      & ZAR  & 12,201 & 2,130  & 464  \\
            \midrule
            Validation & DEP  & 7,414  & 5,044  & 1,611 \\
                      & ZAR  & 2,660  & 443    & 138 \\
            \midrule
            Test       & DEP  & 13,982 & 9,395  & 2,488 \\
                      & ZAR  & 4,990  & 903    &  260  \\
            \bottomrule
        \end{tabular}
        \caption{Statistics on NAIST Text Corpus 1.5}
        \label{tab:ntc-stats}
    \end{minipage}
    \hspace{.01\linewidth}
    \begin{minipage}[b]{.48\linewidth}
        \centering
        \begin{tabular}{lrr}
            \toprule
            \multicolumn{1}{c}{Set of POS tags} & \multicolumn{1}{c}{Ratio (\%)} & \multicolumn{1}{c}{\#Tokens} \\ \midrule
            \masknoun{} & 43.54 & 954,816 \\
            \maskverb{} & 11.34 & 248,777 \\
            \maskparticle{}  & 25.37 & 556,296 \\
            \masksymbol{}  & 11.78 & 258,393 \\
            \maskall{} & 100.00 & 2,192,900 \\
            \bottomrule
        \end{tabular}
        \caption{Number of tokens whose POS tag belongs to a set of POS tags: each number is counted on the training set of NTC 1.5.}
        \label{tab:word-type}
    \end{minipage}
\end{table}

\begin{table}[t]
  \small
  \centering
  \begin{tabular}{@{}ll@{}}
  \toprule
     \textbf{Configurations}         &   \textbf{Values} \\ \midrule
     RNN Cell     &   Bi-GRU~\cite{matsubayashi:2018:COLING}, $k=10$ \\
     Optimizer              &   Adam~\cite{kingma:2015:ICLR} ($\beta_{1}=0.9, \beta_{2}=0.98, \epsilon=1\times10^{-8}$)  \\
     Learning Rate Schedule &   Same as described in Appendix A of Matsubayashi and Inui~\shortcite{matsubayashi:2018:COLING} \\
     Number of Epochs       &   150   \\
     Stopping Criterion     &   Same as described in Appendix A of Matsubayashi and Inui~\shortcite{matsubayashi:2018:COLING} \\ 
     Gradient Clipping      &   1.0  \\
     BERT Fine-tuning       &   None: All BERT parameters are fixed \\
     \bottomrule
  \end{tabular}
\caption{List of hyper-parameters}
\label{tab:hyper-param}
\end{table}

In this section, we first investigate the effectiveness of \textit{\mask{}-based augmentation} by comparing it with the baseline CDA (Section~\ref{subsec:CDA}) on the validation set (Section~\ref{subsec:experiment-mask-based-augmentation}).
We then investigate the relationships between performance gain and each POS category to be replaced with some masking probability to find the best configuration of \textit{linguistically-controlled masking} over the validation set (Section~\ref{subsec:experiment-1}).
Finally, we compare the best configured model with the other existing models over a test set (Section~\ref{subsec:experiment-2}).

\subsection{Experimental Configuration}
\label{subsec:experimental-configuration}
The NAIST Text Corpus (NTC) 1.5~\cite{iida2010annotation,iida2017naist} with the standard dataset splits proposed by Taira et al.~\shortcite{taira2008} was used for the experiments.
NTC 1.5 is a benchmark dataset commonly adopted by previous studies~\cite{ouchi2017,matsubayashi:2018:COLING,omori2019}.
Table~\ref{tab:ntc-stats} presents the number of instances in NTC 1.5, and Table~\ref{tab:word-type} presents the number of tokens whose POS tag appears in the NTC 1.5 training set.

We used both the modified sentence $\bm{X}^{\prime}$ and the original sentence $\bm{X}$ in a 1:1 ratio for training.
Each model was trained using 10 different random seeds.
The POS tags from the NTC gold standard tags were used in the experiments.
The $\argmax$ function was employed as $\mathrm{F}$ in Equation~(\ref{eqn:sampling}), which denotes the function for selecting the most probable word from the probability distribution.
The average \fscore{}-scores were reported for both ZAR and DEP.\footnote{While we focused on ZAR, our results included DEP in accordance with a previous research on this task.}
The BERT~\cite{devlin:2019:NAACL} pretrained using the Japanese Wikipedia~\cite{shibata2019} was employed as the MLM.
The BERT parameters were kept fixed throughout the experiments.
The target predicate was never masked.\footnote{Masking the target predicate reduced the \fscore{} scores in our preliminary experiment.}
Table~\ref{tab:hyper-param} shows the set of employed hyper-parameters.

\subsection{Effectiveness of \mask{}-based Augmentation}
\label{subsec:experiment-mask-based-augmentation}

\begin{figure}[t]
    \def\@captype{table}
    \begin{minipage}[b]{0.48\textwidth}
        \small
        \tabcolsep 4pt
        \centering
        \begin{tabular}{lcccc}
            \toprule
            Method                 & ZAR \fscore{}  & DEP \fscore{}  & ALL \fscore{} \\
            \midrule
            \textsc{Baseline}      & 64.08          & 92.82          & 87.43           $\pm$0.14 \\
            \textsc{Baseline (2x)} & 63.55          & 92.75          & 87.24           $\pm$0.17 \\
            \textsc{CDA}           & 64.30          & 92.79          & 87.42           $\pm$0.18 \\
            \textsc{Masking}       & \textbf{64.89} & \textbf{92.94} & \textbf{87.64}  $\pm$0.09 \\
            \bottomrule
        \end{tabular}
        \tblcaption{\fscore{} scores on the NTC 1.5 \textbf{validation set}. \textbf{Bold} values indicate best results in the same column.}
        \label{tab:validation-scores-compare-with-replace}
    \end{minipage}
    \hfill
    \begin{minipage}[b]{0.46\textwidth}
        \centering
        \includegraphics[width=\hsize]{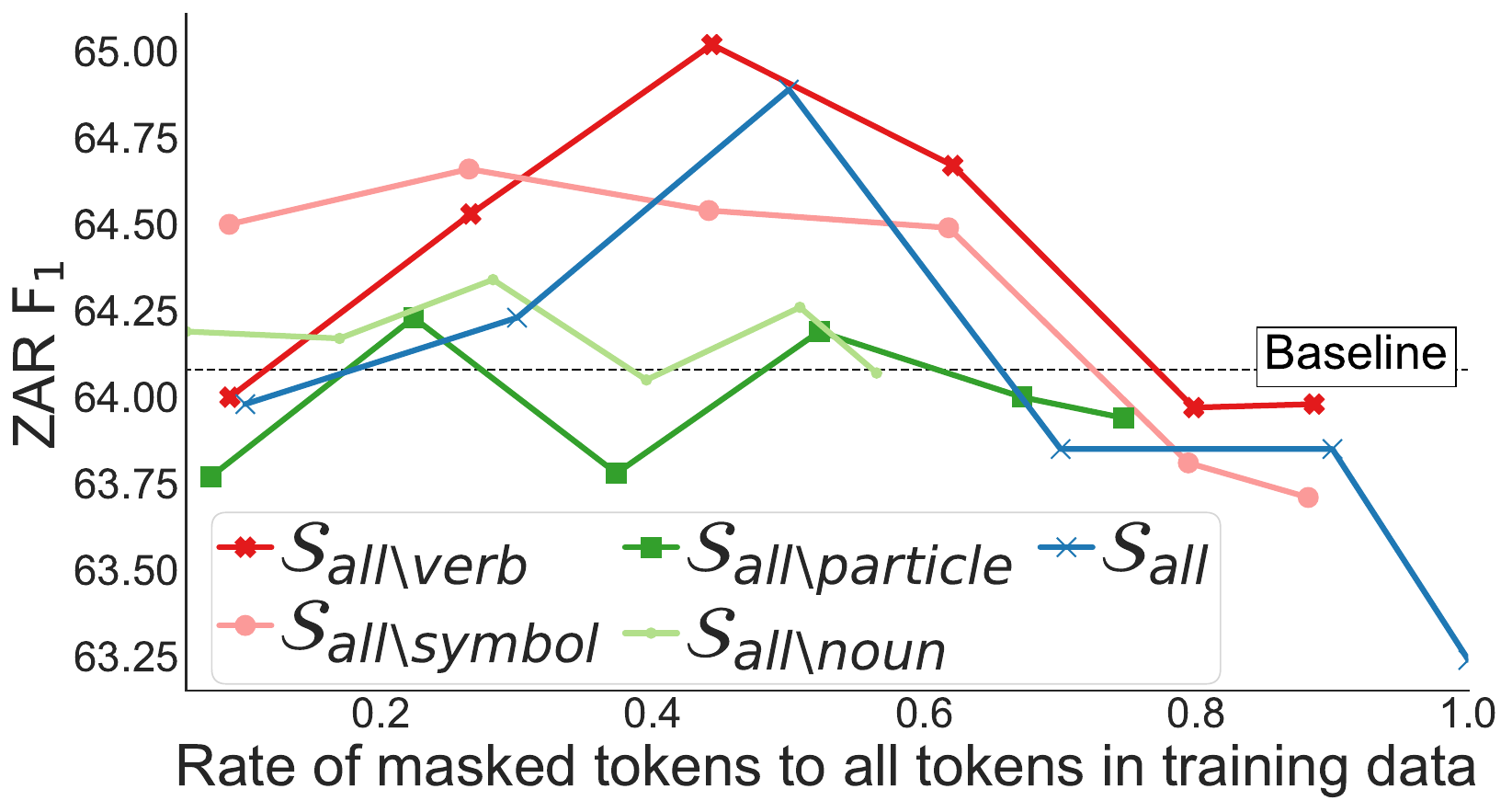}
        \caption{The effect of changing masking probability $\alpha$ on ZAR \fscore{}}
        \label{fig:validation-pos-masking}
    \end{minipage}
\end{figure}

We investigated the effectiveness of the \mask{}-based augmentation, where all tokens were considered as targets for replacement.
The best masking probability $\alpha$ was identified from \{0.1, 0.3, 0.5, 0.7, 0.9, 1.0\}.
The masking positions were fixed for each epoch to reduce the computational cost.
Hereafter, we refer to the proposed method as \textsc{Masking}, which was compared to the base CDA.
In the \textsc{CDA}, all \mask{} tokens were filled simultaneously.\footnote{One may fill a single \mask{} token at a time. However, we found that the method of simultaneously filling all \mask{} tokens is empirically superior. The details are presented in Appendix~\ref{sec:word-replacement}.} 
We also report herein on the performance of the \textsc{Baseline} (i.e., \baseline{}) with double amount of training data, namely, \textsc{Baseline (2x)}.
\textsc{Baseline (2x)} aims to conduct a fair comparison using the same amount of labeled data as those used by \textsc{CDA} and \textsc{Masking}.

Table~\ref{tab:validation-scores-compare-with-replace} shows the results.
\textsc{Masking} achieved the best F$_1$ scores for all the categories.
Note that \textsc{Baseline (2x)} did not outperform \textsc{Baseline}, indicates that the improvement was achieved due to \textsc{Masking} rather than the increased amount of the same training instances.
\textsc{Masking} also outperformed the base CDA, illustrating that our simplification applied to the base CDA contributed to the prediction performance and the computational efficiency.

\subsection{Effectiveness of Linguistically-controlled Masking}
\label{subsec:experiment-1}

We varied a target POS tagset for the masking and the masking probability to investigate the relationships between performance gain and each POS category.
We left each one out of all the POS categories and treated the rest as targets for replacement.
Let \masknoverb{} denote that all POS categories, except for the verb, are targets for replacement (i.e., ``all-but-verb masking,'').
We will refer to the others in the same manner.
The models were trained with every possible combination of the following settings: (i) POS category \{\masknonoun{}, \masknoverb{}, \masknoparticle{}, \masknosymbol{}, \maskall{}\}\footnote{We selected the POS occupying more than 10\% of the corpus.} and (ii) masking probability $\alpha$: \{0.1, 0.3, 0.5, 0.7, 0.9, 1.0\}.\footnote{1.0 denotes masking all tokens of the target POS.}

Figure~\ref{fig:validation-pos-masking} shows the relationships between ZAR \fscore{} and the number of masked tokens using each POS-controlled masking.
The horizontal axis represents the proportion of masked tokens to all tokens in the training data.
Here, all-but-verb masking \masknoverb{} tended to achieve the best results, and all-but-symbol masking \masknosymbol{} was on par or slightly better than \maskall{}.
While these three outperformed \textsc{Baseline} in most cases, the others (i.e., \masknonoun{} and \masknoparticle{}) did not.
These performance differences were also observed under the condition of the same number of masked tokens.
Controlling masked tokens by their POS category affected the performance.
Moreover, a relatively high masking probability of approximately $0.5$ tends to allow the realization of a better performance.

Table~\ref{tab:validation-scores} shows the result of the model achieving the highest ZAR \fscore{} for each POS-controlled masking.
Here, we also trained masking models with a single target POS category (e.g., \maskverb{}).
The results show that all-but-verb masking \masknoverb{} achieved a better ZAR \fscore{} than \textsc{Baseline}.
By contrast, only-verb masking \maskverb{} did not improve the performance compared with \textsc{Baseline}.
These results imply that verbs should not be the targets for replacement.
In this way, the POS category-based analysis and observation are facilitated by our linguistically-controlled masking.
A more detailed analysis is provided in Section~\ref{sec:analysis}.

\subsection{Comparison with the Existing Models}
\label{subsec:experiment-2}

\begin{table}[t]
  \centering
  \begin{tabular}{lcccc}
  \toprule
  Method                  & ZAR \fscore{}  & DEP \fscore{}  & ALL \fscore{}  \\
  \midrule
  \textsc{Baseline}       & 64.08          & 92.82          & 87.43 $\pm$0.14 \\
  \maskall                & 64.89          & 92.94          & 87.64 $\pm$0.09 \\
  \masknoun               & 64.62          & 92.87          & 87.53 $\pm$0.09 \\
  \maskverb               & 64.15          & 92.78          & 87.35 $\pm$0.09 \\
  \maskparticle           & 64.31          & 92.82          & 87.43 $\pm$0.19 \\
  \masksymbol             & 64.12          & 92.74          & 87.29 $\pm$0.16 \\
  \masknonoun             & 64.34          & 92.85          & 87.44 $\pm$0.16 \\
  \masknoverb             & \textbf{65.02} & 92.96          & 87.67 $\pm$0.11 \\
  \masknoparticle         & 64.23          & 92.83          & 87.44 $\pm$0.15 \\
  \masknosymbol           & 64.66          & \textbf{92.98} & 87.59 $\pm$0.19 \\
  \masknoverbnosymbol     & 64.81          & 92.97          & \textbf{87.72} $\pm$0.16 \\
  \bottomrule
  \end{tabular}
  \caption{
      \fscore{} scores for each POS-controlled masking on the NTC 1.5 \textbf{validation set}. The method names denote a set of target POS for masking. \textbf{Bold} values indicate best results in the same column.
  }
  \label{tab:validation-scores}
\end{table}

\begin{table}[t]
    \small
    \centering
    \begin{tabular}{l|cccc|c|cc}
    \toprule
    \multicolumn{1}{l|}{} & \multicolumn{4}{c|}{ZAR} & \multicolumn{1}{c|}{DEP} &     &    \\
    Method                & ALL & NOM & ACC & DAT    & ALL                      & ALL & SD \\ 
    \midrule
    \midrule
    \multicolumn{8}{c}{\textsc{Single Model}} \\
    \midrule
    Matsubayashi and Inui~\shortcite{matsubayashi:2018:COLING}
        & 55.55          & 57.99          & 48.9          & 23          & 90.26          & 83.94 & $\pm$0.12   \\
    Omori and Komachi~\shortcite{omori2019}
        & 53.50          & 56.37          & 45.36         & 8.70        & 90.15          & 83.82 & $\pm$0.10 \\
    \midrule
    \textsc{Baseline}\ourresult{}
        & 63.89          & 66.45          & 57.2          & 27          & 92.24          & 86.85 & $\pm$0.11  \\
    \textsc{CDA}\ourresult{}
        & 63.87          & 66.16          & \textbf{58.5} & \textbf{29} & 92.27          & 86.84 & $\pm$0.19  \\
    \textsc{Masking} (\masknoverb)\ourresult{}  
        & \textbf{64.15} & \textbf{66.60} & 57.9          & \textbf{29} & \textbf{92.46} & \textbf{86.98} & $\pm$0.13  \\
    \midrule
    \midrule
    \multicolumn{8}{c}{\textsc{Ensemble Model}} \\
    \midrule
    Matsubayashi and Inui~\shortcite{matsubayashi:2018:COLING}\ensemble{}
        & 58.07          & 60.21          & 52.5          & 26          & 91.26          & 85.34 & - \\ 
    \textsc{Baseline}\ourresult{}\ensemble{}
        & 66.45          & 68.91          & 60.1          & \textbf{30} & 93.07          & 88.06 & - \\
    \textsc{Masking} (\masknoverb)\ourresult{}\ensemble{}
        & \textbf{67.35} & \textbf{69.85} & \textbf{61.2} &  28         & \textbf{93.32} & \textbf{88.43} & - \\ 
    \bottomrule
    \end{tabular}
    \caption{
        \fscore{} scores on the NTC 1.5 \textbf{test set}. 
        \textbf{Bold} value indicate the best results in the same column group. 
        $\dagger$: result from our experiment. 
        $*$: ensemble models.
        The improvement of \textsc{Masking} over the \textsc{Baseline} is statistically significant in both overall F1 score and ZAR F1 score ($p < 0.05$) with a permutation test.
    }
    \label{tab:test-scores}
\end{table}

We compared the model built by all-but-verb masking \masknoverb{}, which achieved the best ZAR \fscore{} on the validation set (Table~\ref{tab:validation-scores}), with state-of-the-art models~\cite{matsubayashi:2018:COLING,omori2019} on the test set.
Table~\ref{tab:test-scores} shows the results.
We refer to our method of combining \textit{\mask{}-based augmentation} and \textit{linguistically-controlled masking} in the optimal setting as \textsc{Masking} (\masknoverb{}).
Our \textsc{Baseline} already outperformed the state-of-the-art results by a large margin.
When new techniques are only evaluated on very basic models, determining how much (if any) improvement will carry over to stronger models can be difficult~\cite{denkowski:2017,suzuki:2018}.
In contrast to this concern, \textsc{Masking} (\masknoverb{}) consistently achieved the statistically significant improvements over the strong \textsc{Baseline} in both single and ensemble-model settings.
This solid finding is likely to be generalizable to other strong models.
On the contrary, the ZAR \fscore{} of the \textsc{CDA} was on par with that of \textsc{Baseline}, which shows that the symbolic replacement of tokens is not effective.

\section{Analysis of Augmented Data}
\label{sec:analysis}
\begin{figure}[t]
    \begin{minipage}[b]{0.495\linewidth}
        \centering
        \includegraphics[width=\hsize]{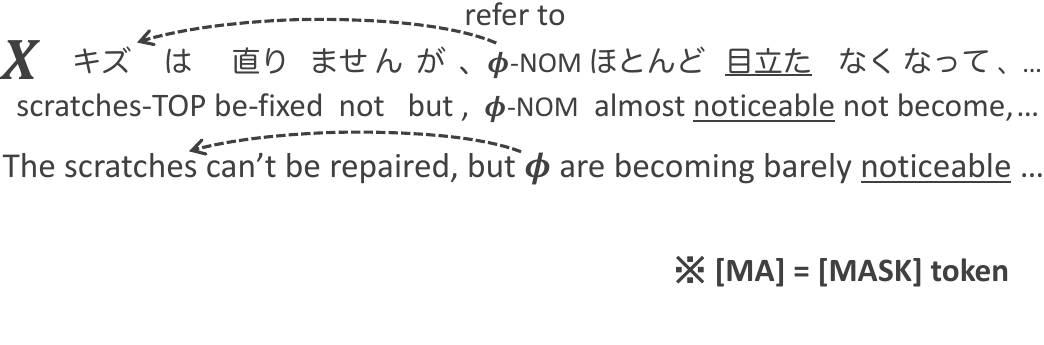}
    \end{minipage}
    \hfill
    \begin{minipage}[b]{0.50\linewidth}
        \centering
        \includegraphics[width=\hsize]{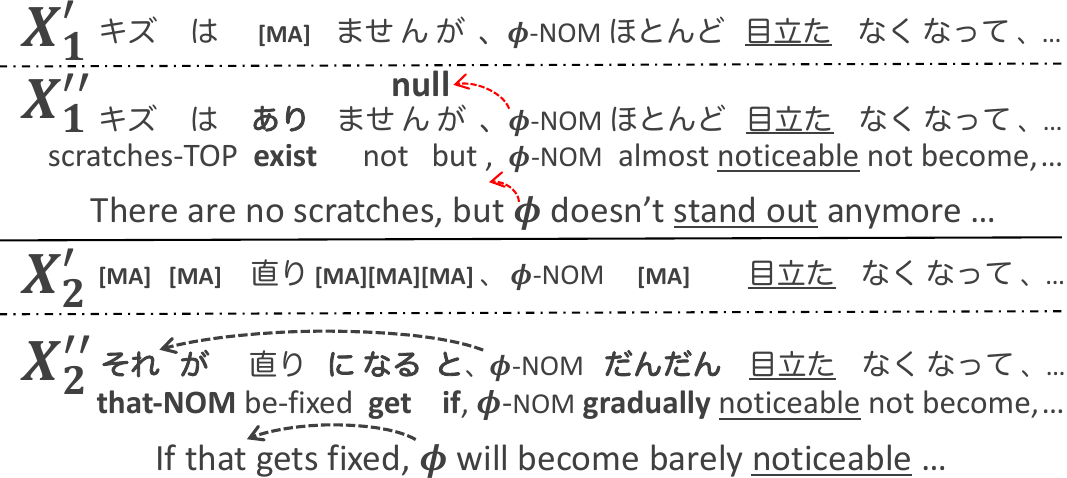}
    \end{minipage}
    \caption{Examples of pseudo data created by data augmentation.
            $\bm{X}^{\prime}$ indicates the pseudo data in \textsc{Masking}.
            $\bm{X}^{\prime\prime}$ indicates the pseudo data in \textsc{CDA}.
            $\bm{X}_1^{\prime}$ and $\bm{X}_1^{\prime\prime}$ are output of \maskverb{}.
            $\bm{X}_2^{\prime}$ and $\bm{X}_2^{\prime\prime}$ are output of \masknoverb{}.
            }
    \label{fig:example-for-analysis}
\end{figure}

In this section, we analyze the actual instances augmented by \textit{linguistically-controlled masking}.
The model built by all-but-verb masking \masknoverb{} achieved the best ZAR \fscore{}; hence, we further investigated the negative effect of verb replacement.
We did this by observing the actual instances produced by \masknoverb{} and verb-only masking \maskverb{}.
Figure~\ref{fig:example-for-analysis} depicts the three types of sentences of these two models, that is, the original ones $\bm{X}$, the masked ones $\bm{X}^{\prime}$, and those with some symbolically replaced tokens $\bm{X}^{\prime\prime}$.
Note that although our method did not use any symbolic substitutions $\bm{X}^{\prime\prime}$ for predicting the of zero anaphora (Section~\ref{sec:mask-based-augmentation}), they were produced only for the analysis.\footnote{We computed the probability distribution using the representation of each \mask{} in the same manner as a standard CDA and produced the token with the highest probability.}
Looking at the symbolic token generated from each mask provides us an intuition on the feature vector of each mask.
For example, if the token \textit{exist} is generated from a mask, we can interpret that its corresponding vector approximately represents the meaning of \textit{exist}.

In the original sentence $\bm{X}$, the nominative argument of the target predicate \textit{noticeable} was realized as a zero pronoun $\phi$, and its antecedent was \textit{The scratches}.
In the sentence generated by \maskverb{}, the symbolic token \textit{exist} filled the \mask{} ($\bm{X}_1^{\prime\prime}$), and the semantic structure of the sentence, especially the original relation between the two predicates (verbs) \textit{be fixed} and \textit{noticeable}, was changed.
The antecedent can be interpreted as \textit{null}, which was changed from the original one \textit{The scratches}.
In contrast, in the sentence generated by \masknoverb{} ($\bm{X}_2^{\prime\prime}$), the antecedent of the zero pronoun was not changed, even though half of the tokens in the sentence has been replaced.
The antecedent was \textit{that}, which aligned with the original token (position), \textit{scratches}.
The original relation between the two predicates, \textit{be fixed} and \textit{noticeable}, still remained.
In addition, the pretrained MLM generated the tokens surrounding these predicates, such that the context drastically changed from the original one \textit{the scratches can't be repaired, but \dots} to \textit{if that gets fixed, \dots}.
Such a contextual variation plays an important role for improving the robustness and the coverage of a ZAR model, even if the meaning is slightly unnatural.
This can be a reason for the results in Section~\ref{sec:linguistically-controlled masking}, indicating that the setting of a high masking probability (approximately $0.5$) and all-but-verb masking \masknoverb{} achieves a better performance.

\section{Discussion}
We tackled a data scarcity problem on the Japanese ZAR through data augmentation.
We specifically improved the CDA for the ZAR from the quantitative and qualitative aspects: \textit{\mask{}-based augmentation} (Section~\ref{sec:mask-based-augmentation}) and \textit{linguistically-controlled masking} (Section~\ref{sec:linguistically-controlled masking}).

\noindent\textbf{Summary of Empirical Results}\hspace{3mm}
We obtained the following results from the extensive experiments (Section~\ref{sec:experiment}):
\begin{itemize}
\setlength{\parskip}{0cm} 
\setlength{\itemsep}{0cm}
    \item Effectiveness of \textit{\mask{}-based augmentation}:
    efficiently enables integrating a gigantic MLM and achieves statistically significant improvements over a strong baseline and standard CDA (Table~\ref{tab:validation-scores-compare-with-replace});
    \item Effects of masking probability: a relatively high masking probability leads to a  performance gain (Figure~\ref{fig:validation-pos-masking});
    \item Advantage of \textit{linguistically-controlled masking}: this enables controlling which types of tokens to be replaced. Verbs should not be targets for replacement (Table~\ref{tab:validation-scores}).
\end{itemize}

\noindent\textbf{Suggestions from In-depth Analysis}\hspace{3mm}
The analysis on the actual augmented data (Section~\ref{sec:analysis}) provided the future ZAR research with two suggestions:
\begin{itemize}
\setlength{\parskip}{0cm} 
\setlength{\itemsep}{0cm}
    \item Keep the original semantic structure, especially the relation between the predicates (verbs); otherwise, the original antecedent of a zero pronoun is likely to be changed.
    \item Generate contextual variations relatively far from an original sentence by adjusting the masking probability while keeping its original semantic structure.
\end{itemize}

\noindent\textbf{Future Work}\hspace{3mm}
We found herein the interesting phenomenon of \textit{change of antecedents} and analyzed it by observing several actual instances.
One interesting direction of the future research is the quantitative investigation of this phenomenon and the revelation of the relationships between the change of antecedents and the types of tokens to be replaced.
We controlled the token replacement positions by POS tags.
The other possible method of determining these positions is to use a syntactic structure, such as dependency relations, and relative distances from the target predicate.
We plan to explore such linguistic features to control the replacement.

\bibliographystyle{coling}
\bibliography{coling2020}

\appendix
\newpage
\onecolumn

\section{Word Replacement Approaches}
\label{sec:word-replacement}
This section compares alternative approaches of the \textsc{CDA} method described in Section~\ref{subsec:CDA}.
We compare several different procedures for filling \mask{} tokens using the pretrained MLM.
Specifically, we consider two aspects: \textbf{(i) how many \mask{} tokens to fill simultaneously} and \textbf{(ii) how to determine a word symbol in each time step}.

\begin{figure}[tb]
    \begin{minipage}[b]{0.65\linewidth}
        \centering
        \includegraphics[width=\hsize]{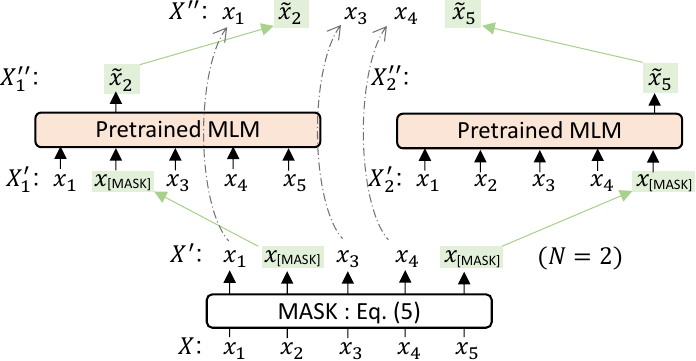}
        \subcaption{SINGLE}
        \label{fig:appendix-single}
    \end{minipage}
    \hspace{.01\linewidth}
    \begin{minipage}[b]{0.30\linewidth}
        \centering
        \includegraphics[width=\hsize]{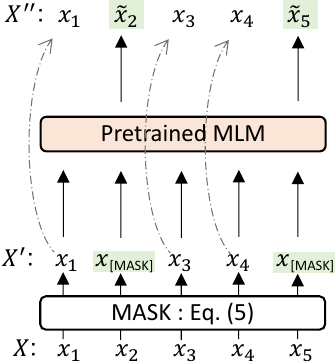}
        \subcaption{MULTI}
        \label{fig:appendix-multi}
    \end{minipage}
    \caption{Overview of two replacement approaches.}
    \label{fig:app-replacement}
\end{figure}

For \textbf{(i) how many \mask{} tokens to fill simultaneously}, we compare the following two methods:
\begin{itemize}
  \item \textsc{Single}: 
    The overview of this method is presented in Figure~\ref{fig:appendix-single}. Suppose that a masked sentence $\bm{X}^{\prime}$ contains $N$ \mask{} tokens, we first create $(\bm{X}_{1}^{\prime},\dots,\bm{X}_{N}^{\prime})$ from $\bm{X}^{\prime}$. Here, each $\bm{X}_{n}^{\prime}$ contains a single \mask{} token and the rest of tokens are from the original sentence $\bm{X}$. Note that the position of the \mask{} token is unique for each $\bm{X}_{n}^{\prime}$. We then feed $(\bm{X}_{1}^{\prime},\dots,\bm{X}_{N}^{\prime})$ to MLM independently and fill each \mask{} token with another token to obtain $(\bm{X}_{1}^{\prime\prime},\dots,\bm{X}_{N}^{\prime\prime})$. Finally, we merge the outputs $(\bm{X}_{1}^{\prime\prime},\dots,\bm{X}_{N}^{\prime\prime})$ to construct a symbolically modified sentence $\bm{X}^{\prime\prime}$.
  \item \textsc{Multi}: 
    The overview of this method is presented in Figure~\ref{fig:appendix-multi}. Given a masked sentence $\bm{X}^{\prime}$, we feed $\bm{X}^{\prime}$ to MLM and fill all \mask{} tokens simultaneously. The output of MLM is a symbolically modified sentence $\bm{X}^{\prime\prime}$.
\end{itemize}

For \textbf{(ii) how to determine a word symbol in each time step}, we compare the following two methods:
\begin{itemize}
  \item \textsc{Sample}: Following Wang and Cho~\shortcite{wang:2019:bert}, we determine the output symbol by sampling a word from the probability distribution over vocabulary that is computed by pretrained MLM.
  \item \textsc{Argmax}: From the probability distribution over the vocabulary, we determine the output symbol by choosing the token with the highest probability.
\end{itemize}

\begin{table}[tb]
  \centering
  \small
  \begin{tabular}{lrrrr}
  \toprule
  Method & ZAR \fscore{} & DEP \fscore{}  & ALL \fscore{} & SD \\
  \midrule
  \textsc{Single}+\textsc{Argmax}    &  63.59  & 92.62  &  87.16 &  $\pm$0.12 \\
  \textsc{Single}+\textsc{Sample}    &  63.58  & 92.55  &  87.14 &  $\pm$0.16 \\
  \textsc{Multi}+\textsc{Argmax}     &  \textbf{64.47}  & \textbf{92.70}  &  \textbf{87.37} &  $\pm$0.15 \\
  \textsc{Multi}+\textsc{Sample}     &  63.18  & 92.25  &  86.84 &  $\pm$0.21 \\
  \bottomrule
  \end{tabular}
  \caption{\fscore{} scores on the NTC 1.5 \textbf{validation set}. \textbf{Bold} values indicate the best results in the same column group.}
  \label{tab:replacement-scores}
\end{table}

Table \ref{tab:replacement-scores} shows the result of the combinations of these two aspects.
In this experiment, the conditions of the POS category are identical across all models with the setting of \masknoverbnosymbol, which achieves the best overall \fscore{} in Table~\ref{tab:validation-scores}.
The Table \ref{tab:replacement-scores} presents that the combination of \textsc{Multi} and \textsc{Argmax}, namely, \textsc{Multi}+\textsc{Argmax} achieves the best performance.
Thus, we used the \textsc{Multi}+\textsc{Argmax} setting for \textsc{CDA} in the experiment of Section~\ref{subsec:experiment-2}.

\end{document}